\documentclass[5pt]{article}

\usepackage[letterpaper]{geometry}
\usepackage{spconf,amsmath,epsfig}
\usepackage{enumitem}

\usepackage{fancyhdr}
\begin{document}\sloppy

\def\x{{\mathbf x}}
\def\L{{\cal L}}

\title{Unsupervised Incremental Learning of Deep Descriptors \\ from Video Streams}
%
\name{Federico Pernici and Alberto Del Bimbo}
\address{MICC -- University of Florence\\
federico.pernici@unifi.it, alberto.delbimbo@unifi.it}
%
%
%

\maketitle

\begin{abstract}
  We present a novel unsupervised method for face identity learning from video sequences. The method exploits the ResNet deep network for face detection and VGGface fc7 face descriptors together with a smart learning mechanism that exploits the temporal coherence of visual data in video streams. We present a novel feature matching solution based on Reverse Nearest Neighbour and a feature forgetting strategy that supports incremental learning with memory size control, while time progresses. It is shown that the proposed learning procedure is asymptotically stable and can be effectively applied to relevant applications like multiple face tracking.
\end{abstract}
%
%
\section{Introduction}
\label{sec:introduction}
Visual data is massive and is growing faster than our ability to store and index it, nurtured by the diffusion and widespread use of social platforms. Their fundamental role in advancing object representation, object recognition and scene classification research have been undoubtedly assessed by the achievements of Deep Learning \cite{krizhevsky2012imagenet}. However, the cost of supervision, as necessary for effective training, remains the most critical fact for the applicability of such learning methods.  Efforts to collect large quantities of annotated images, such as ImageNet \cite{deng2009imagenet}, Microsoft coco \cite{lin2014microsoft}, Megaface \cite{kemelmacher2016megaface} and Visual Genome \cite{krishnavisualgenome}, while having had an important role in advancing object recognition, don't have the necessary scalability and are hard to be extended or replicated.  Semi or unsupervised Deep Learning from image data still remains hard to achieve.

An attracting alternative would be to learn the object appearance from video streams with no supervision, both exploiting the large quantity of video available in the Internet and the fact that adjacent video frames contain semantically similar information, so providing the variety of conditions in which an object can be framed, and therefore a comprehensive representation of its appearance.
According to this, tracking a subject in the video could, at least in principle, support a sort of unsupervised incremental learning of its appearance. This would avoid or reduce the cost of annotation as time itself would provide a form of weak supervision.  However, this solution is not free of problems. On the one hand, parameter re-learning of Deep Networks, to adequately incorporate the new information without catastrophic interference, is still an open challenge \cite{li2016learning,rusu2016progressive}, especially when re-learning should be done in real time, while tracking. On the other hand, classic object tracking has substantially divergent goals from continuous incremental learning. While in tracking the object appearance is learned only for detecting the object in the next frame (the past information is gradually forgotten), continuous incremental learning would require that all the past visual information of the object is collected in a comprehensive representation. This requires that tracking does not drift in the presence of occlusions or appearance changes, and at the same time memory overflow is avoided by retaining only the most distinctive descriptors. Finally, incremental learning should be asymptotically stable in order to converge to an univocal representation.
\\
\indent
In this paper, we present unsupervised learning of subject identities from video streams that exploits the ResNet deep network \cite{hu2016finding} to detect faces in consecutive images and fc7 deep descriptor of VGGface  \cite{Parkhi15} for face representation, together with a smart incremental learning mechanism that collects such descriptors and distills the most distinctive ones of them, in order to provide a sufficiently complete representation of the individual identities and at the same time avoid memory overflow. It is shown that under reasonable assumptions our learning procedure is asymptotically stable. The incremental learning mechanism has been inspired by the research on long-term tracking described in \cite{alienpami}. In that system distinctive SIFT local features of a target were detected in each frame and their descriptors were all collected in a template with random forgetting of the past. This ultimately allowed to learn a temporally-updated collection of local features that was useful to track the target in the long term with almost no drifting. 
\\
\indent
In the following, in Section 2, we cite a few works that have been of inspiration for our work. In Section 3, we highlight our contributions and expounded the approach in detail and finally, in Section 4, some experimental results are given.

\section{Related Work}

One key point of the method is the exploitation of video temporal coherence as a form of weak supervision. This idea was suggested by \cite{wang2015unsupervised}, but was essentially applied with some success to predict future frames with unsupervised feature learning,  \cite{vondrick2015anticipating} among the most notable experiments. 

Inclusion of a memory mechanisms in learning is another key feature of our approach. Works  on parameters re-learning on domains that have some temporal coherence, have used reinforcement learning, \cite{mnih2013playing} \cite{schaul2015prioritized} among the most recent ones. They typically store the past experience in a replay memory with some priority and sample mini-batches for training. This makes it possible to break the temporal correlations by mixing more and less recent experiences. More recently, Neural Turing Machine architectures have been proposed in \cite{santoro2016one} and \cite{graves2014neural} that implement an augmented memory to quickly encode and retrieve new information. These architectures have the ability to rapidly bind never-before-seen information after a single presentation via an external memory module. However, in these cases, training data are still provided supervisedly and the methods don't scale with massive video streams. 

Finally, another relevant research subject to our learning setting is long-term object tracking \cite{LTDT2014}. Only a few works on tracking have reported drift-free results on on very long video sequences ( \cite{kalalCVPR2010, MedioniCVPR2011, alienpami, hua2014occlusion, Hong_2015_CVPR} among the few), and only few of them have provided convincing evidence on the possibility of  incremental learning strategies that are asymptotically stable \cite{kalalCVPR2010} \cite{alienpami}. However, all of these works only address tracking and perform incremental learning just to  detect the target in the next frame. 
\\

\section{The proposed approach }

\begin{figure}
\centering    \includegraphics[width=0.99\columnwidth]{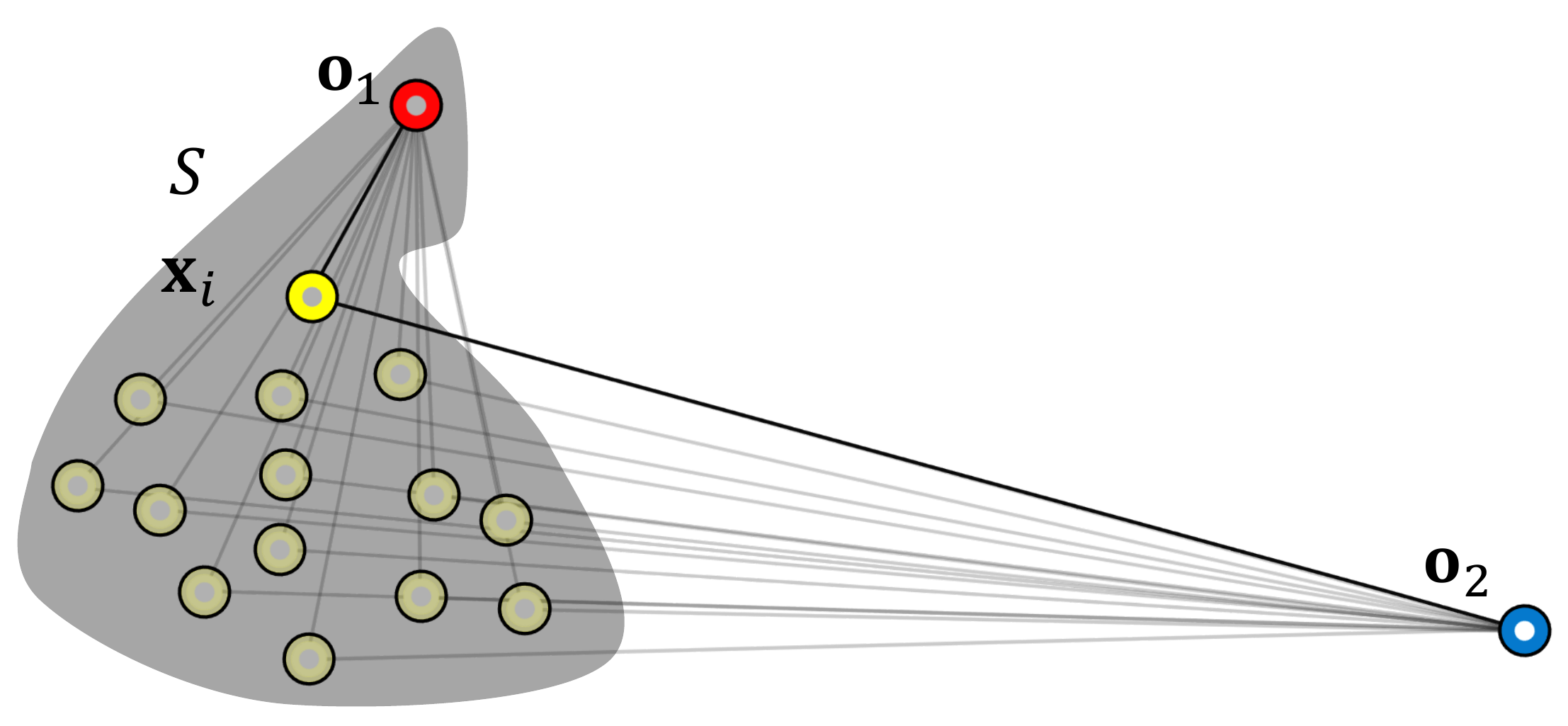}
\caption{Reverse Nearest Neighbor for a repeated temporal visual structure with the distance ratio criterion. All elements $\mathbf{x}_i$ match with $\mathbf{o}_1$, for clarity only one of them is highlighted. }
\label{fig_RNN}
\end{figure}

In our system, fc7 face descriptors of VGGface deep network are computed on face windows detected by ResNet and stored in a memory module as:
\begin{eqnarray}
\mathcal{M}(t) =  \{  (\mathbf{x}_i,\text{Id}_i, {e}_i)  \}^{N(t)}_{i=1}
\end{eqnarray}
where $\mathbf{x}_i$ is the descriptor computed at the fc7layer, $\text{Id}_i$ is the object identity (an incremental number), ${e}_i$ is the eligibility factor (discussed in the following) and $N(t)$ is the number of descriptors at time $t$ in the memory module. 

As video frames are observed, new faces are detected and their descriptors are matched with those already in the memory. Each  newly observed $\mathbf{o}_i$ descriptor, will be assigned with the object identity of its closest neighbour. Unmatched descriptors of the faces in the incoming frame are stored in the memory module with a new $\text{Id}$. They ideally represent faces of new individuals that have not been observed already and will eventually appear in the following frames. While matching these descriptors with those already in the memory permits to track each individual face properly through the video frames, two distinct problems must however be solved in order to collect all the distinguishing descriptors and learn a comprehensive identity of each observed subject. They respectively are concerned with matching in consecutive frames and control of the memory module. These are separately addressed in the following subsections.

\subsection{Reverse Nearest Neighbour matching} 
\label{Template  updating}

While tracking the faces in consecutive frames, it is likely that the face of the same individual will have little differences from one frame to the following. In this case, highly similar descriptors will be stored in the memory and quickly a new face descriptor of the same individual will have comparable distances to the nearest and the second nearest descriptor already in the memory. In this case, the Nearest Neighbor (NN) classifier distance-ratio \cite{lowe} does not work properly and matching cannot be assessed.  We solved this problem  by performing descriptor matching according to Reverse Nearest Neighbour (ReNN) \cite{Korn2000}:  
\begin{eqnarray}
\mathcal{M}^\star = \Big \{  (\mathbf{x}_i,\text{Id}_i, e_i) \in \mathcal{M}(t) \: | \: \frac{||\mathbf{x}_i-1\mathrm{NN}_{I_t}(\mathbf{x}_i)||}{||\mathbf{x}_i-2\mathrm{NN}_{I_t}(\mathbf{x}_i)||} < \bar{\rho}, \, \Big  \}
\label{RNNtemplate}
\end{eqnarray}
where $\bar{\rho}$ is the distance ratio threshold, $\mathbf{x}_i$ is a face descriptor in the memory module and $1\mathrm{NN}_{I_t}(\mathbf{x}_i)$ and $2\mathrm{NN}_{I_t}(\mathbf{x}_i)$ are respectively its nearest and second nearest neighbor face descriptor in the incoming frame $I_t$.
\\
\indent
Fig.~\ref{fig_RNN} shows the effects of this change of perspective: here two new observations are detected (two distinct faces, respectively marked as $\mathbf{o}_1$ and $\mathbf{o}_2$). They both have distance ratio close to 1 to the nearest $\mathbf{x}_{i}$s in the memory (the dots in the grey region). Therefore both their matchings are undecidable. Differently from NN, ReNN is able to correctly detect the nearest descriptor for each new descriptor in the incoming frame. In fact, with ReNN, the roles of $\mathbf{x}_i$ and $\mathbf{o}_i$ are exchanged and the distance ratio is computed between each $\mathbf{x}_i$ and the $\mathbf{o}_{i}$ as shown in figure for one of the $\mathbf{x}_i$s (the yellow dot is associated to the newly observed red dot). 
Due to the fact that with ReNN a large number of descriptors (those accumulated in the memory module) is matched against a relatively small set of descriptors (those observed in the current image), calculation of the ratio between distances could be computationally expensive if sorting is applied to the entire set. However, minimum distances can be efficiently obtained by performing twice a linear search, with parallel implementation on GPU. 

\subsection{Memory size control} 
Descriptors that have been matched according to ReNN ideally represent different appearances of a same subject face. However, collecting these descriptors indefinitely could quickly determine memory overload. To detect redundant descriptors and discard them appropriately, we defined a dimensionless quantity $e_i$ referred to as \emph{eligibility}. This is set to $e_i = 1$ as a descriptor is entered in the memory module and hence decreased at each match with a newly observed descriptor, proportionally to the distance ratio: 
\begin{equation}
e_{i}(t+1) = \eta_{i} \, e_{i}(t). 		\label{eligibility}
\end{equation}
Eligibility allows to take into account both spatial redundancy (close descriptors) and temporal updating (only the most recent matched descriptors are retained). In fact, as the eligibility $e_{i}$ of a face descriptor $\mathbf{x}_i$ in the memory drops below a given threshold $\bar{e}$ (that happens after a number of matches), that descriptor is removed from the memory module: 
\begin{equation}
\mbox{if} \: (e_{i} < \bar{e}) \: \mbox{then} \: \mathcal{M}(t+1) = \mathcal{M}(t) \setminus \{ (\mathbf{x}_i,\text{Id}_i, {e}_i) \}.
\label{eligibilityThreshold}
\end{equation}
The value $\eta_{i}$ is computed according to:
\begin{equation}
	\eta_{i} = \frac{1}{\bar{\rho}}
	\bigg[ \frac{d^{1}_i}	
				{d^{2}_i} \bigg]^\alpha,			
    \label{eq_squaredratio}
\end{equation} 
where $d^{1}_i$ and $d^{2}_i$ are respectively the distances between $\mathbf{x}_i$ and its first and second nearest neighbour $\mathbf{o}_i$, the value $\bar{\rho}$ is the distance-ratio threshold of Eq.~\ref{RNNtemplate}, used for normalization and $\alpha$ emphasizes the effect of the distance-ratio. 

Some of the features collected in the memory will never obtain matches. This effect is largely due to scene occluders or object features with very low repeatability. In the long run such features may waste critical space in the memory buffer. They are handled by considering a max time lapse in which a descriptor has not not been matched, after which the descriptor is discarded. The threshold can be set reasonably large to avoid deletion of rare but useful descriptors.

\subsection{Asymptotic stability}

Under the assumption that descriptors are sufficiently distinctive (as in the case of VGGface fc7 descriptors), the incremental learning procedure described above stabilizes asymptotically around the probability density function of the descriptors of each individual subject face. A key element which guarantees such theoretical asymptotic stability is that the ReNN distance ratio is always below 1. In fact, it is easy to demonstrate that the updating rule of Eq.~\ref{eligibility} is a contraction and converges to its unique fixed point 0 according to the Contraction Mapping theorem (Banach fixed-point theorem).

\begin{figure}[h]
\centering	
\includegraphics[width=0.35\textwidth]{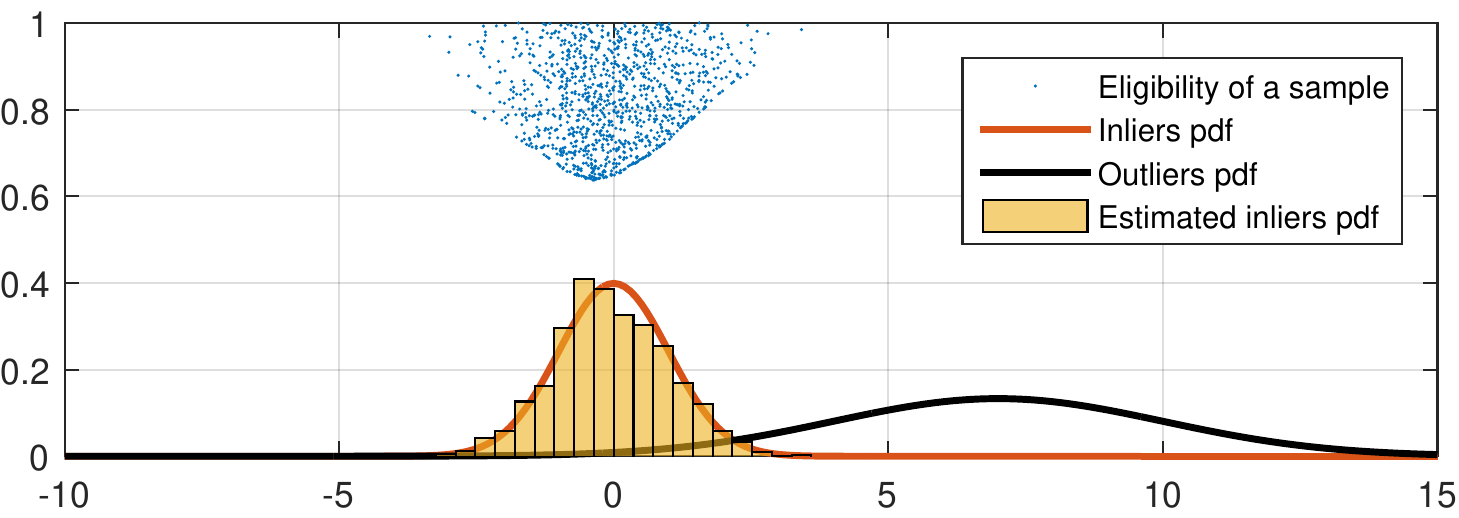} 	
\includegraphics[width=0.35\textwidth]{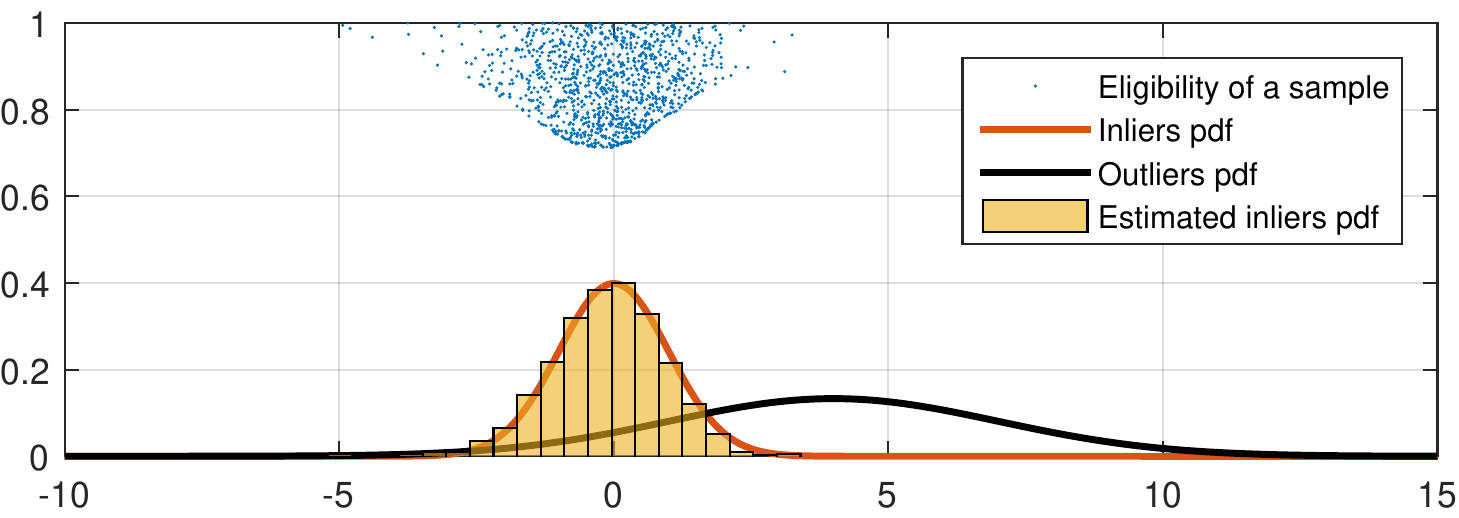} 	
\includegraphics[width=0.35\textwidth]{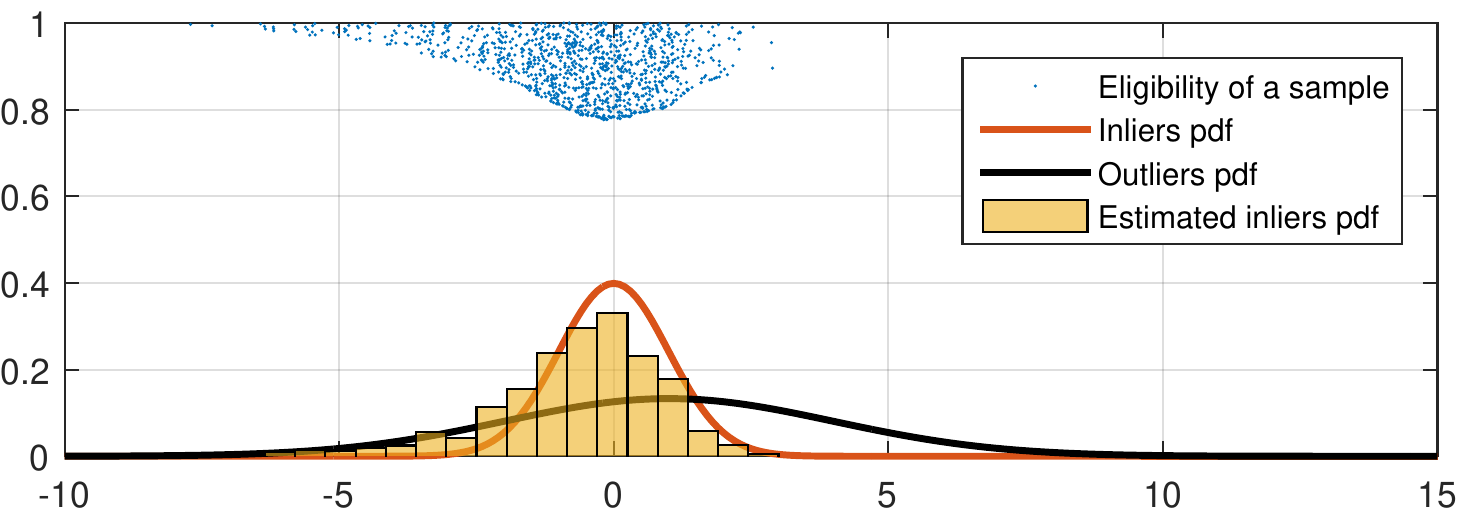}
\caption{Asymptotic stability of incremental learning of a face identity in a sample sequence}.
\label{asymptotic}
\end{figure}

The asymptotic stability of the method and its robustness to erroneous matches is illustrated in Fig.~\ref{asymptotic} with a simple one-dimensional case. Two patterns of synthetic descriptors, respectively modeling the inliers and the outliers of some object identity, are generated by two distinct 1D Gaussian distributions. 
The learning method was ran for 1000 iterations for three different configurations of the two distributions. 
The blue points represent the eligibility of the descriptors of the object instances detected. The red curve is the inliers pdf. The black curve is the outlier pdf. The histogram in yellow represents the distribution of the inliers as incrementally learned by the system. The inliers pdf models the case of true positive detections (the object descriptors have little differences from each other as subsequent frames have little motion and context changes). The outliers pdf instead models the case of false positives detected (the standard deviation of the descriptors is therefore larger). Mismatches might therefore corrupt the inlier distribution. 

The three figures represent distinct cases in which the outlier pdf is progressively overlapping the inlier pdf. When the outliers are sufficiently far from the inliers, that is the descriptors are distinctive (top), the density of the eligibility closely follows the density of the data distribution (samples that are close to the pdf mode are matched more frequently and their eligibility decreases accordingly). As the outliers pdf gets closer to the inliers pdf, the ReNN matching mechanism and the memory control mechanism still keep the learned inlier pdf close to the ground truth pdf (medium, bottom).  Close outliers only determine some little bias of the distribution. Despite of the fact that part of the symmetric and peaked shape of the eligibility is lost, the mode and standard deviation of the distribution does not substantially change. 

\section{Learning from a video stream}

Evaluation of performance of our solution for unsupervised learning from tracking in video streams presents some difficulties. While for supervised learning cross-validation is typically used, and train and test datasets are availble that permit direct comparison between learning methods, unfortunately, for our unsupervised incremental learning, cross-validation cannot be applied. In fact it is impossible to know in advance the exemplars of the moving objects and their variations \cite{ACMsurveyConcept}.  
\\
\indent
We provide here two experiments that respectively show performance of our learning mechanism in terms of precision/recall at different learning rounds and the capability of the learning mechanism to support a multiple face tracking application. 

In the first experiment, we collected a number of YouTube videos (185) that contain the face of a well known public person, namely the former US President Barack Obama, with high probability (retrieved from the query: "Barack Obama"), for a total of 58 hours (6.264.000 frames). These videos eventually include the face of President Obama at different times, in either indoor or outdoor settings, under different illumination and occlusion conditions. The original resolution of Video data was $640\times360$. Video frames were re-sized to $320 \times 240$ pixels and images to $320 \times 240$ pixel. These videos were used as snapshots for training only.  

Hence we issued the same query on Google Images, and collected 5000 images of Barack Obama. Additional 5000 images that did not contain the subject were included in the image dataset. We assumed that the videos and images sets were someway correlated and that video data were sufficient to learn some views of the appearances of the face of President Obama. All the images were manually annotated and used during test. The image test set was split in two subsets of 5000 images each. One (Subset A) was used to improve the quality of learning and the other (Subset B) to derive a measure of performance of the learning mechanism.

\begin{figure}
\centering
\includegraphics[width=0.74\columnwidth]{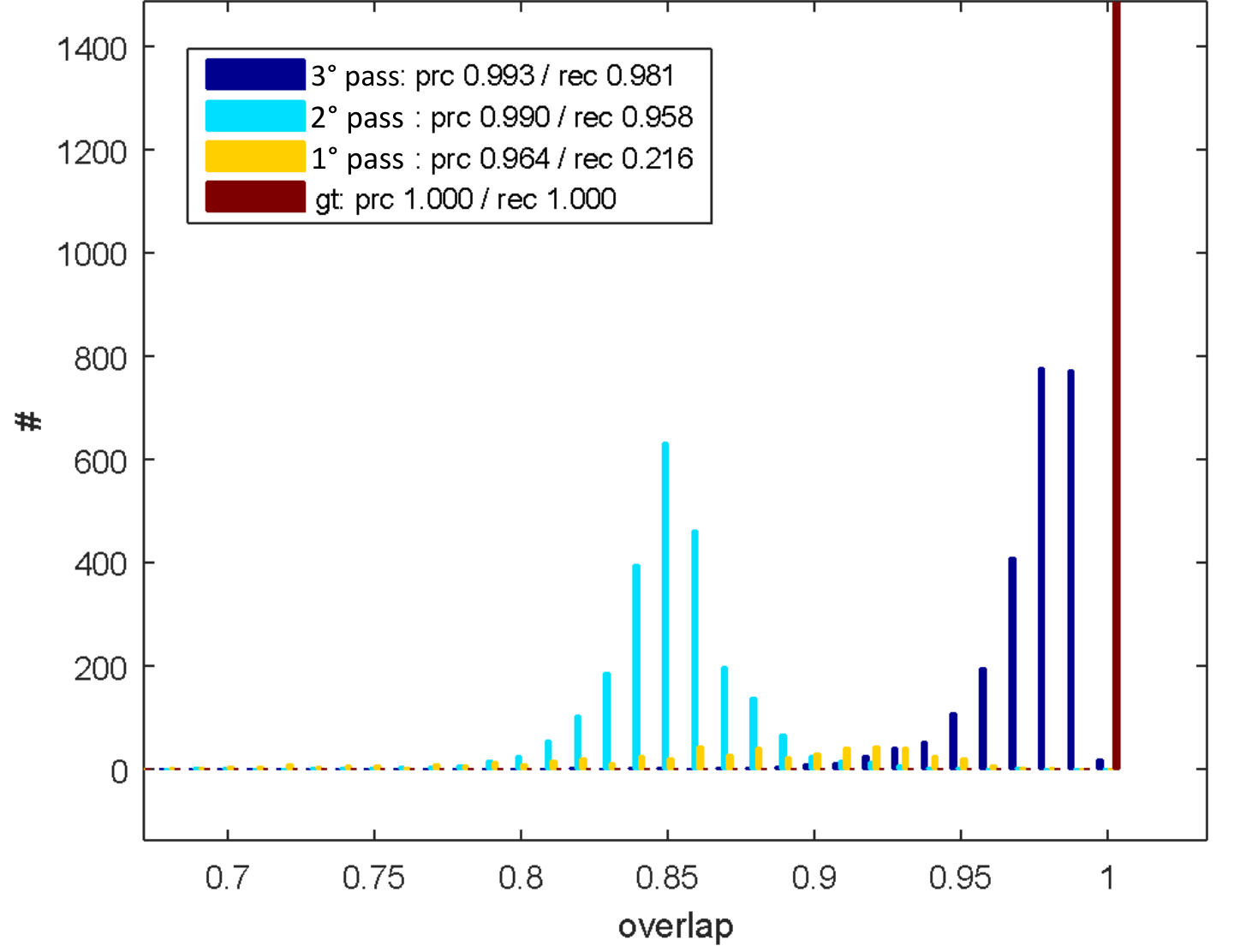}	
\caption{Histogram of matches at different overlaps for different passes over image subset A, and corresponding Precision and recall values computed on image Subset B. Recall is computed considering 0.5 matching threshold.  }
\label{fig_histCount}
\end{figure}

The learner, trained on the videos, was applied to image subset A repeatedly, for several passes, so that at each pass new descriptors of the face of President Obama were included in the memory module and the eligibilities of the descriptors in the memory were updated so that the learner incrementally learns more of the Obama's face.  At each pass the learner was hence run onto test image subset B (for which a ground truth is known) and the Precision and Recall were evaluated. Fig.\ref{fig_histCount} explains the internals of the learning mechanism.  Three different steps are shown with the histograms of the matches at different overlaps between the bounding box predicted by the detector and the ground truth. The blue histogram (final pass) reveals a distribution close to the ground truth (the single brown bar). in the white box, the values of Precision and Recall at the same passes are shown. It is clearly visible the convergence of the learning mechanism. As a new pass over the image subset A is performed, the performance increases substantially, nearly up to the optimal result.

\begin{figure*}
\centering
\includegraphics[width=0.9\textwidth]{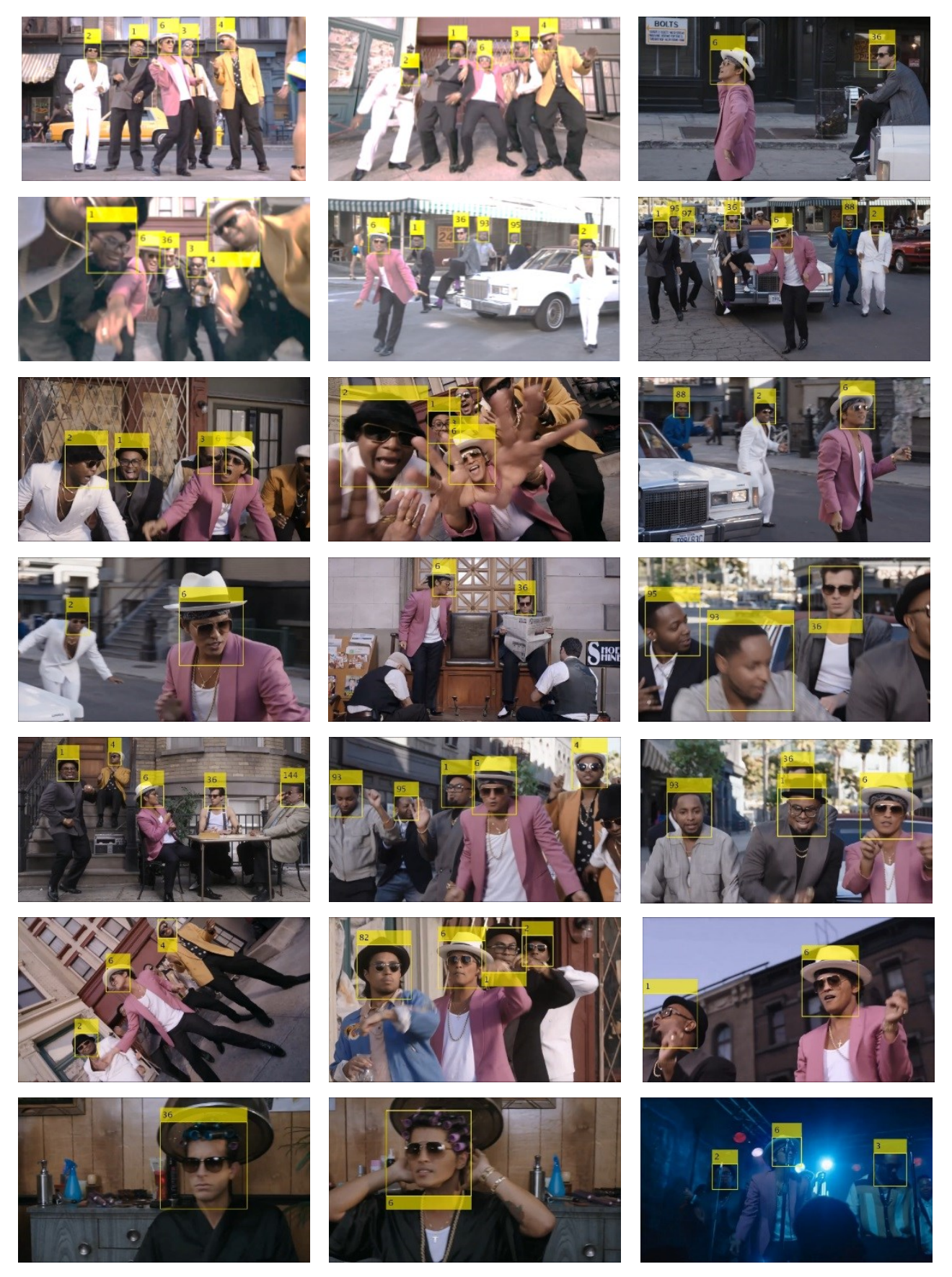}
\caption{The unsupervised learning method applied to Multiple Face Tracking. Selected frames from the \emph{BrunoMars} video sequence with the superimposed estimated identities are shown.}
\label{brunoMarsFrames}
\end{figure*}

In the second experiment we directly applied the incremental learner system to a video sequence and verified (qualitatively) its capability to learn the distinct identities of the faces unsupervisedly. Fig.\ref{brunoMarsFrames} shows a few frames of the video clip of "Bruno Mars" from the dataset \cite{zhang2016tracking} and the learned identities (the numbers superimposed to the faces detected). The video clip was composed putting together different shots taken in different places, where the faces of the characters appear in different orientations, illumination conditions, and poses. In some cases the same individual (see f.e. individual number 6) appears sensibly diverse (with and without hat). Nonetheless it can be observed that the learning mechanism is sufficiently capable to exploit the temporal coherence between frames and collect the distinguishing descriptors of each individual so that its identity is not lost.

\section{Conclusion}
In this paper we exploited  deep network based face detection and fc7 VGGface descriptor coupled with a novel learning mechanism that learns face identities from video sequences unsupervisedly, exploiting the temporal coherence of video frames.
The proposed method is simple, theoretically sound, asymptotically stable and follows the cumulative and convergent nature of human learning.
\\

\section*{Acknowledgment}  
This research is based upon work supported in part by the Office of the Director of National Intelligence (ODNI), Intelligence Advanced Research Projects Activity (IARPA), via IARPA contract number 2014-14071600011. The views and conclusions contained herein are those of the authors and should not be interpreted as necessarily representing the official policies or endorsements, either expressed or implied, of ODNI, IARPA, or the U.S. Government. The U.S. Government is authorized to reproduce and distribute reprints for Governmental purpose notwithstanding any copyright annotation thereon.

\small
\bibliographystyle{IEEEbib}
\bibliography{camera-ready_icme2017template}

\end{document}